\documentclass[10pt,twocolumn,letterpaper]{article}

\usepackage{tabularx}
\usepackage[pagenumbers]{cvpr} % To force page numbers, e.g. for an arXiv version

\definecolor{cvprblue}{rgb}{0.21,0.49,0.74}
\usepackage[pagebackref,breaklinks,colorlinks,allcolors=cvprblue]{hyperref}

\def\paperID{32} % *** Enter the Paper ID here
\def\confName{CVPR}
\def\confYear{2026}

\title{A Large-Scale Study on the Accuracy vs Cost Trade-offs of \\ Training and Evaluation Settings in Fine-Grained Image Recognition}

\author{
    \textbf{Edwin Arkel Rios\textsuperscript{\dag}, Augusto Christian Surya\textsuperscript{\ddag}, Oswin Gosal\textsuperscript{\ddag}, Fernando Mikael\textsuperscript{\ddag},}\\
    \textbf{Mary Madeline Nicole\textsuperscript{\ddag}, Kisoon Jang\textsuperscript{\ddag}, Bo-Cheng Lai\textsuperscript{\dag}, Min-Chun Hu\textsuperscript{\ddag} } \\
    \textsuperscript{\dag}\textit{National Yang Ming Chiao Tung University, Taiwan}, \textsuperscript{\ddag}\textit{National Tsing Hua University, Taiwan} \\
}

\begin{document}
\maketitle
\begin{abstract}
Prior work on fine-grained image recognition (FGIR) has established the importance of the backbone selection, but has neglected the accuracy-vs-cost trade-offs under different training and evaluation settings. In this work we conduct a large-scale study with over 2000 experiments across 6 training and evaluation settings, 9 pretrained backbones, and 17 datasets. Preliminary observations on the effectiveness of data augmentation for fine-grained training motivate us to extend Counterfactual Attention Learning (CAL), a state-of-the-art method based on data-aware cropping and masking augmentations, with cross-image discriminative region mixing augmentation. We also propose an efficient evaluation-only variant that maintains competitive accuracy while reducing inference costs by forfeiting the forward pass on discriminative crops that is normally used by CAL and similar FGIR methods. Our results show that data-aware augmentations during training only can enable a model to achieve excellent accuracy even without crops, significantly reducing inference costs. To support future research we share our code and checkpoints at: \url{https://github.com/arkel23/FGIR-Backbones}
\end{abstract}

\section{Introduction}
\label{sec_introduction}

\begin{figure*}[!htb]
    \centering
    \includegraphics[width=0.75\textwidth]{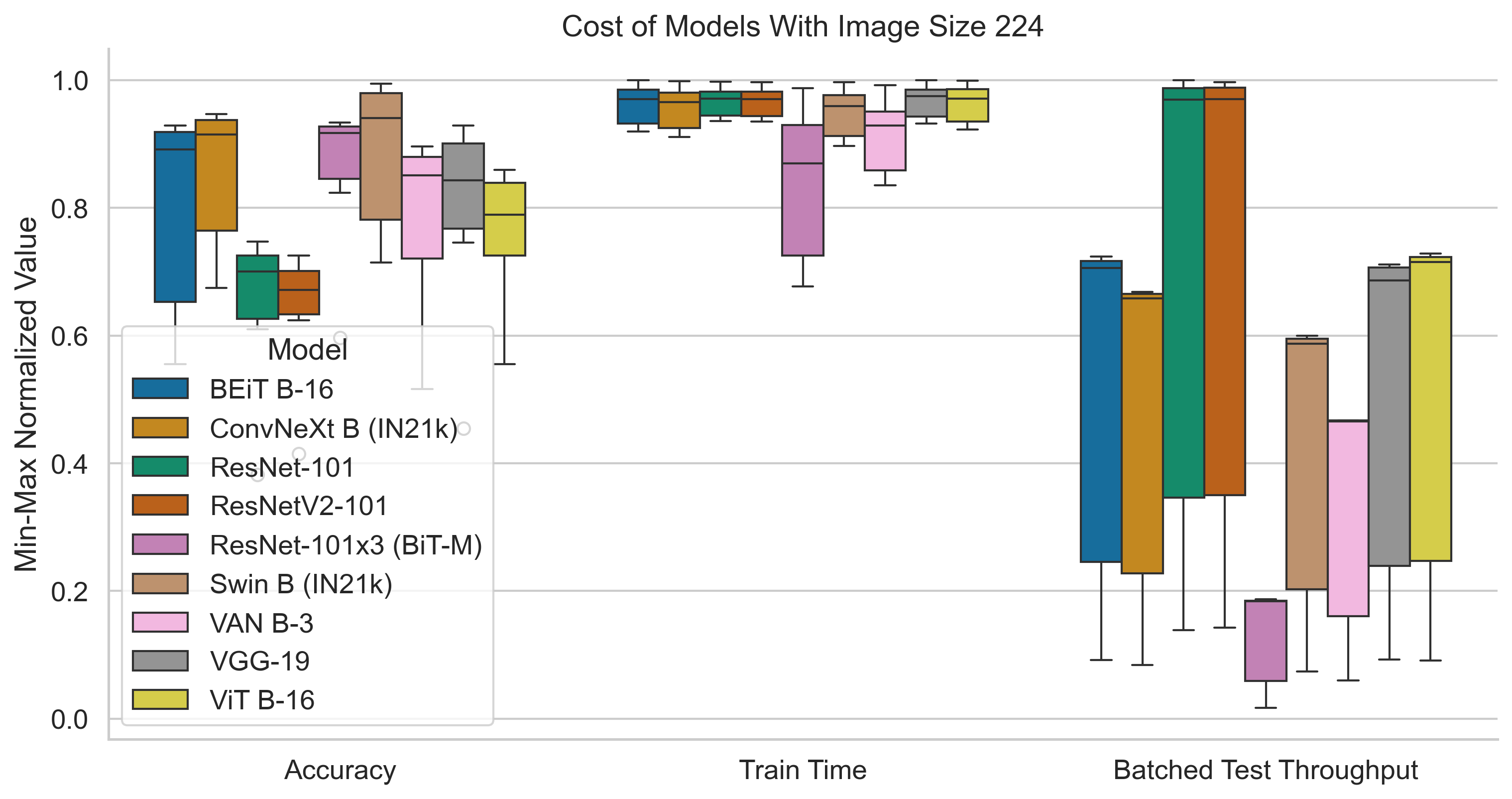}
    \caption{Comparison of 9 Backbones across 13 Datasets. Accuracy as Performance, Train Time \& Inference Throughput as Cost.}
    \label{fig_main_plot}
    \vspace{-0.2cm}
\end{figure*}

Fine-grained image recognition (FGIR) tasks—such as distinguishing bird species or car models—require identifying subtle visual differences between closely related categories. While previous works on FGIR focused on modules for discriminative feature selection \cite{zhang_part-based_2014, zheng_learning_2017, hu_see_2019, rao_counterfactual_2021}, recent studies \cite{ye_image_2024} highlight that the choice of backbone architecture is a critical decision in FGIR; not only the backbone dictates downstream accuracy, but it is a major factor on training and deployment costs \cite{ye_image_2024} as our experiments show in \cref{fig_main_plot}.

However, performance is also influenced by the choice of training and evaluation settings (\textbf{TrEvS}). Prior work \cite{kornblith_better_2019, goldblum_battle_2023, ye_image_2024, p_which_2024} primarily benchmarks backbones under generic recognition settings and overlooking the impact of FGIR-specific strategies. Therefore, in this work, we systematically study the interactions of different TrEvS—including both generic, frozen or fine-tuned settings and FGIR-specific configurations—on 9 pretrained backbones spanning CNNs and Transformers, and 17 fine-grained datasets with diverse domains and data densities. Our results show that FGIR-specific TrEvS can improve performance up to 60\% relative accuracy, but can also increase computational cost by up to 279\%. Based on this observation we propose a novel method based on counter-factual attention learning (CAL) \cite{rao_counterfactual_2021}. Our method, CALMix, further improves CAL by integrating data-aware cross-image augmentation. To reduce costs, we introduce variations that do not incorporate crops at inference time, CAL-NC and CALMix-NC. This significantly reduces deployment costs while maintaining the benefits of FGIR-aware training. 

Our contribution includes:
\begin{itemize}
    \item We conduct a systematic study of over 2000 experiments—covering 6 training and evaluation settings, 9 backbones, and 17 datasets—on the interaction between backbone and training and evaluation settings. Our results provide practical guidelines for FGIR system design, highlighting that increase in setting complexity can improve accuracy by up to 60\% in relative terms, while incurring up to a 279\% increase in computational cost.
    \item We propose efficient variants of CAL, including \textbf{CAL-NC}, which removes inference-time attention cropping, and \textbf{CALMix}, which further improves performance via stronger data-aware augmentation, and its efficient variant \textbf{CALMix-NC}.
    \item We release the training, evaluation, and benchmarking code, along with over 700 pretrained checkpoints, to support future research in FGIR.
\end{itemize}

\section{Experiment Setup}
\label{sec_methodology}

\begin{figure*}[!htb]
    \centering
    \includegraphics[width=1.0\textwidth]{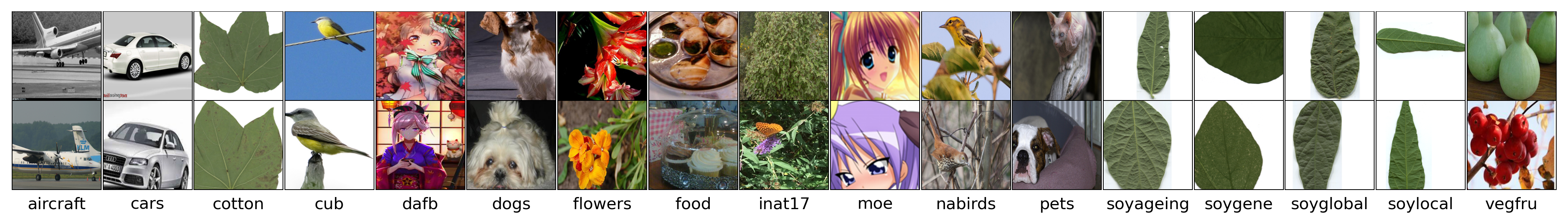}
    \caption{Example images of each dataset used in this experiment.}
    \label{fig_datasets}
\end{figure*}

Our experimental protocol consists of two stages. In the first stage, we perform a hyperparameter search over five learning rates {0.3, 0.1, 0.03, 0.01, 0.003}. For each dataset, we select the best learning rate based on performance on a held-out evaluation subset of the training data, and use it for all subsequent experiments. In the second stage, we conduct multi-seed experiments using the selected learning rate for each dataset to account for stochasticity in deep learning optimization and obtain more robust performance estimates.

\noindent\textbf{Training \& Evaluation Settings.} Prior work \cite{kornblith_better_2019, ye_image_2024, p_which_2024} typically evaluates backbones under fixed configurations, namely \textbf{Frozen (FZ)}, where only the classifier head is trained while the backbone remains fixed, and \textbf{Fine-Tuned (FT)}, where all backbone parameters are updated on the target dataset. However, they neglected FGIR-specific training strategies that can affect performance. To address this, we incorporate \textbf{Counterfactual Attention Learning (CAL)} \cite{rao_counterfactual_2021}, which improves attention by leveraging bilinear attention pooling (BAP) and counterfactual supervision to compare factual and counterfactual predictions, encouraging the model to focus on more discriminative regions.

\noindent\textbf{Improving Efficiency with Train-Only Augmentations and Data-Aware Cross-Image Mixing.} CAL and many FGIR methods \cite{zhang_part-based_2014, zheng_learning_2017, hu_see_2019, hu_rams-trans_2021, rios_global-local_2025} employ a two-stage inference where an attention-like module—BAP in the case of CAL—first predicts discriminative regions based on a global forward pass. These regions are then used to crop the image for a second local forward pass. To separate the impact of these inference attention-crops from the data-aware training, we investigate a variant where CAL is applied only during training, removing attention cropping at inference. We refer to this variant as \textbf{CAL-NC} (CAL No Crops). As shown in Section \ref{sec_result_settings}, CAL-NC reduces inference cost with only a marginal decrease in accuracy. These results suggest that the primary benefit of CAL comes from its data-aware training mechanism. Building on this observation, we explore whether stronger data-aware augmentation can further improve performance. Specifically, in addition to attention cropping and masking, we incorporate cross-image mixing \cite{yun_cutmix_2019, huang_snapmix_2021, zhang_intra-class_2021, zhang_s3mix_2023} by swapping discriminative regions of two images from the same class; we denote this extended training strategy as \textbf{CALMix}. This improves robustness by encouraging region-level feature learning. However, CALMix also involves inference-time attention cropping, increasing deployment costs. To address this, we also introduce an efficient variant without inference crops, \textbf{CALMix-NC}.

\noindent\textbf{Datasets.} We evaluate our approach on 17 fine-grained datasets with diverse domains and data density. To understand the effect of different factors such as backbones, image size or training settings, we initially focused on 4 datasets: FGVC-Aircraft \cite{maji_fine-grained_2013}, CUB-200-2011 \cite{wah_caltech-ucsd_2011}, SoyGene \cite{yu_benchmark_2021}, SoyLocal \cite{yu_benchmark_2021}. We then expand our analysis to 13 additional datasets: Stanford Cars \cite{krause_collecting_2013}, DAFB \cite{gwern_danbooru2021_2015, rios_dafre_2021}, Stanford Dogs \cite{khosla_novel_2012}, Oxford Flowers-102 \cite{nilsback_automated_2008}, Food-101 \cite{yan_discriminative_2022}, iNaturalist \cite{van_horn_inaturalist_2018}, Moe \cite{noauthor_tagged_nodate}, NABirds \cite{van_horn_building_2015}, Oxford-IIIT Pets \cite{parkhi_cats_2012}, Cotton \cite{yu_benchmark_2021}, SoyAgeing \cite{yu_benchmark_2021}, SoyGlobal \cite{yu_benchmark_2021}, and VegFru \cite{hou_vegfru_2017}. Example images of each datasets are shown in Figure \ref{fig_datasets}.

\noindent\textbf{Models.} Our experiments consider 9 backbone architectures spanning both CNNs and Transformers: VGG-19 \cite{simonyan_very_2015}, ResNet-101 (RN-101) \cite{he_deep_2016}, ResNetV2-101 (RNV2-101) \cite{he_identity_2016}, ResNetV2-101x3-BiT (RN-101x3) \cite{kolesnikov_big_2020}, ViT B-16 \cite{dosovitskiy_image_2020}, Swin-B \cite{liu_swin_2021}, ConvNeXt-B \cite{liu_convnet_2022}, and VAN-B3 \cite{guo_visual_2022}.

\noindent\textbf{Evaluation Metrics.} We evaluate performance and cost using the following metrics:
\begin{itemize}
    \item \textbf{Top-1 Accuracy.} The percentage of samples where the predicted class matches the ground-truth label.
    
    \item \textbf{Training Time.} The total time required to complete a full training run, measured in minutes.

    \item \textbf{Batched Test Throughput.} The number of samples processed per second during inference, computed under batched evaluation, as opposed to single-sample.
\end{itemize}

To enable fair comparison across heterogeneous scales, all reported values are normalized using min-max scaling before visualization:
\[
x' = \frac{x - \min(x)}{\max(x) - \min(x)}
\]
This allows consistent visualization of the accuracy–efficiency trade-off within a unified plot.

\begin{figure*}[!htb]
    \centering
    \includegraphics[width=0.8\textwidth]{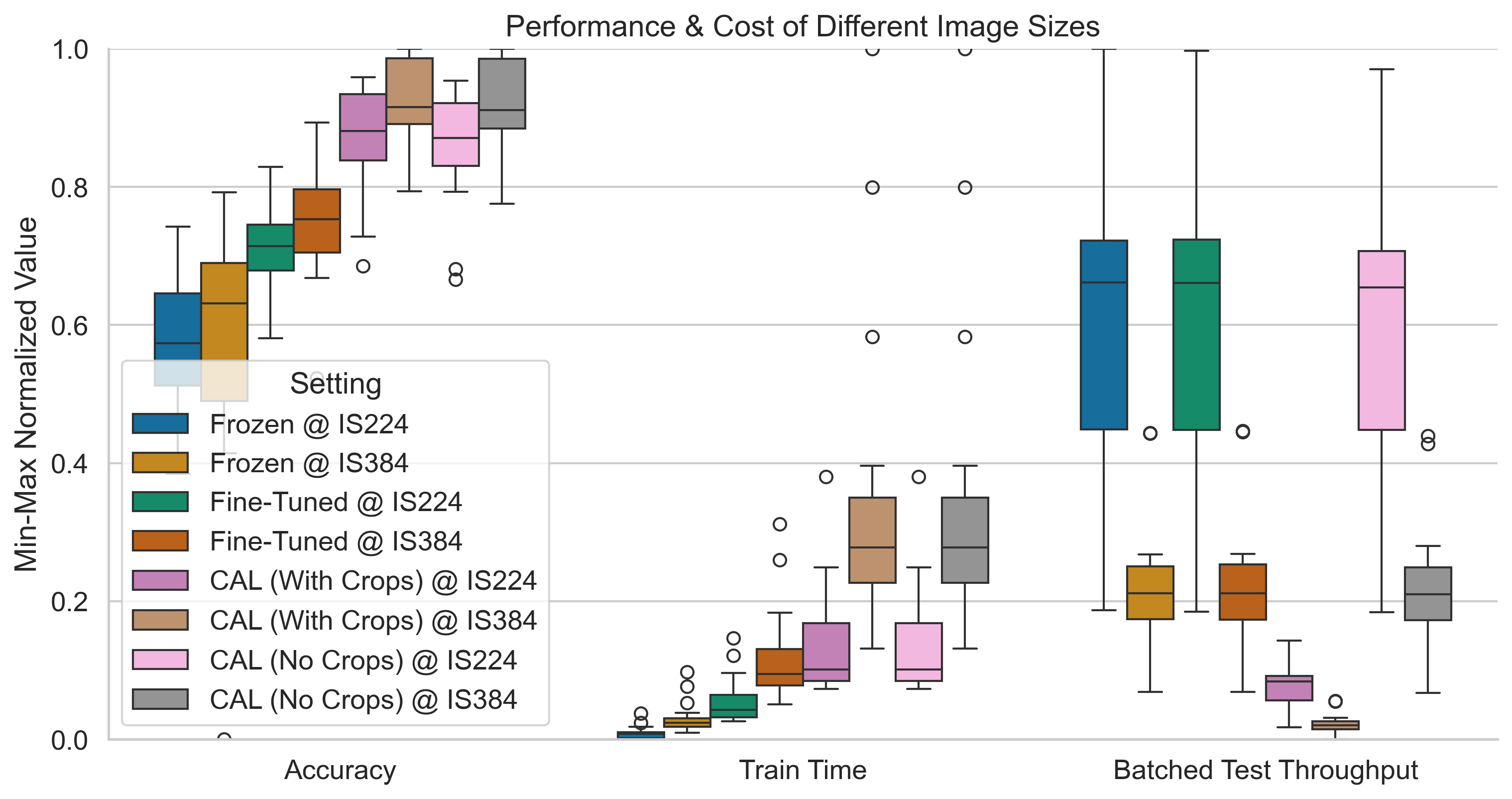}
    \caption{Comparison of FZ, FT, CAL and CAL-NC settings with Image Size 224 (IS224) vs. Image Size 384 (IS384) across 9 Backbones and 4 Datasets. Accuracy as Performance, Train Time \& Inference Throughput as Cost.}
    \label{fig_is_tradeoff}
    \vspace{-0.2cm}
\end{figure*}

\section{Results}
\label{sec_results}
Before presenting detailed analyses, we first summarize the evaluation setup and aggregation strategy. The Backbone, Image Size, and Training \& Evaluation Setting analyses (Figures~\ref{sec_result_backbones}, \ref{sec_result_is}, \ref{sec_result_settings}) aim to study the interaction between model design choices and the performance–cost trade-off, where higher performance generally comes with increased computational cost. The Backbone analysis is conducted across all 17 datasets, while the Image Size and Training Setting analyses are evaluated on the 4 original datasets. Across all three analyses, we report min-max normalized values of accuracy, training time, and inference throughput to ensure fair comparison across heterogeneous scales.

\subsection{Performance \& Cost of Different Backbones}
\label{sec_result_backbones}
As shown in Figure \ref{fig_main_plot}, \textbf{more modern and complex backbone architectures} generally \textbf{incur higher computational cost}, resulting in longer training time and lower inference throughput. Classical CNN-based models such as ResNet-101 exhibit lower computational cost and faster training and inference compared to Transformer-based models such as ViT-B/16, which are more computationally demanding. While more complex architectures often achieve stronger performance, the \textbf{relationship between complexity and accuracy is not strictly monotonic}, and some highly complex models fail to provide gains that justify their additional cost, such as ResNet-101x3 (BiT-M). Among all evaluated backbones, \textbf{Swin-B and ConvNeXt-B achieve the best accuracy–cost trade-off}, consistently outperforming both lighter CNNs and heavier Transformer variants while maintaining moderate computational overhead.

\begin{figure*}[!htb]
    \centering
    \includegraphics[width=0.75\textwidth]{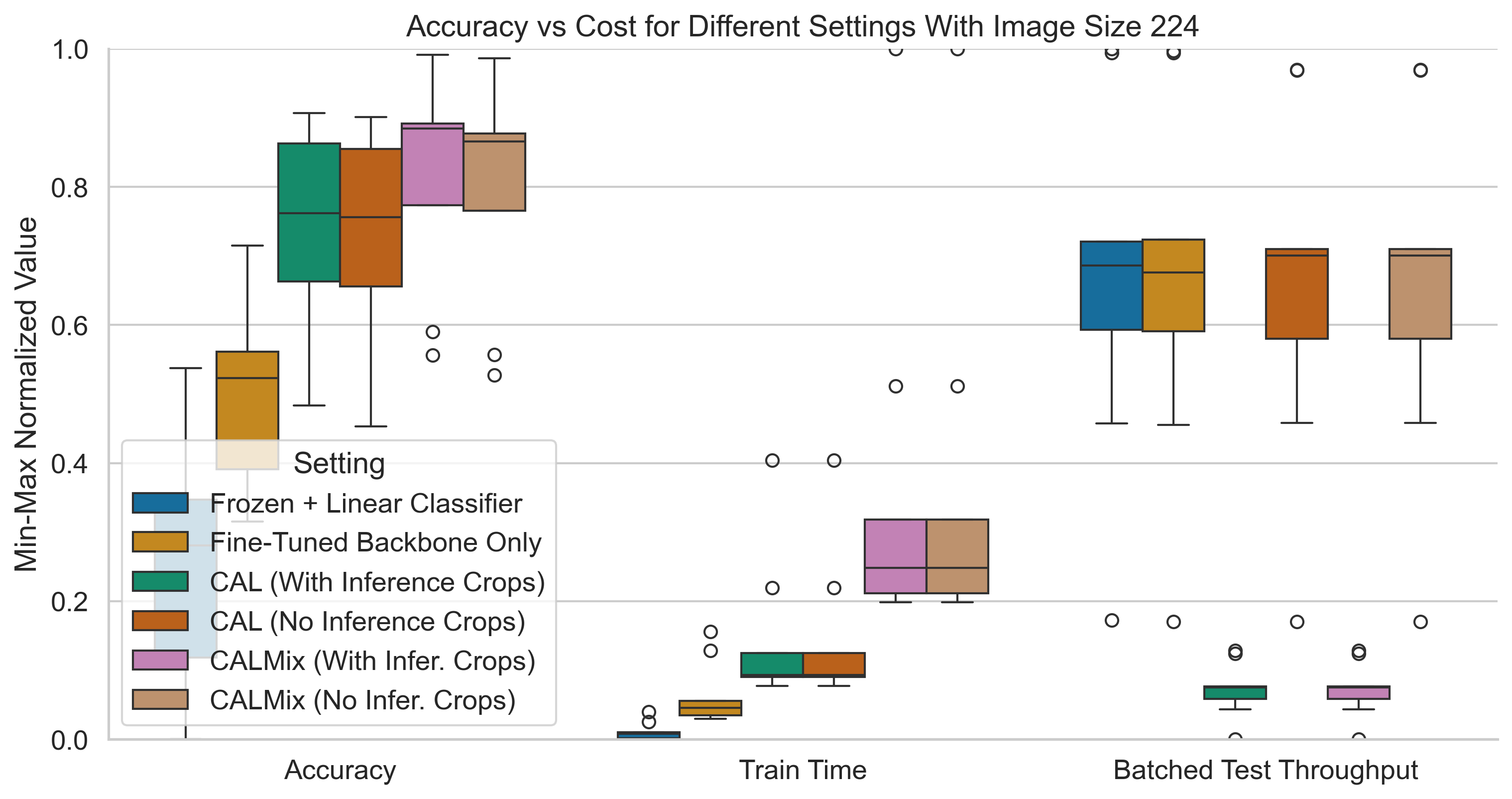}
    \caption{Comparison of 6 Training and Evaluation Settings across 9 Backbones and 4 Datasets. Accuracy as Performance, Train Time \& Inference Throughput as Cost.}
    \label{fig_settings_tradeoff}
\end{figure*}

\subsection{Performance \& Cost of Different Image Sizes}
\label{sec_result_is}
As illustrated in Figure \ref{fig_is_tradeoff}, \textbf{increasing the input resolution} from 224 to 384 \textbf{improves accuracy but incur high computational cost across all training settings} (FZ, FT, CAL and CAL-NC). Higher resolution leads to up to a 217\% relative increase in training time and up to an 82\% relative decrease in inference throughput, while yielding at most a 26\% increase in accuracy. Additionally, \textbf{accuracy gain is not always proportional to computational cost}, and in the frozen (FZ) setting, higher resolution can result in worse performance. These results show the accuracy–cost trade-off when scaling resolution in FGIR, suggesting that \textbf{naive resolution scaling is not an efficient strategy} under computational constraints.

\subsection{Performance \& Cost of Different Settings}
\label{sec_result_settings}
As shown in Figure \ref{fig_settings_tradeoff}, \textbf{the more complex the training \& evaluation settings the more it improves accuracy, but incur higher computational cost}. Transitioning from FZ to FT yields up to a 60\% relative accuracy improvement, while increasing training time by up to 279\%, with negligible change in inference throughput. This suggests that even \textbf{moderate increases in complexity already incur significant cost}. The trend becomes more pronounced with CAL, an FGIR-specific, data-aware training; compared to FT, CAL further improves accuracy by up to 54\%, but increases training time by up to 154\%. Attention-based cropping during inference reduces throughput by up to 90\%. To alleviate this, we propose CAL-NC, which removes inference-time cropping and incurs only up to a 9\% relative accuracy drop, while improving throughput by up to 825\%, restoring efficiency to levels comparable with FZ and FT.

Furthermore, CALMix improves accuracy by up to 38\% over CAL, but increasing training cost by up to 149\%. CALMix-NC maintains an accuracy almost as high as CALMix but reduces inference costs to FT levels. Overall, while FGIR-specific training strategies consistently improve accuracy, they introduce significant computational cost. \textbf{CAL-NC and CALMix-NC provide a strong accuracy–efficiency trade-off}, making them practical choices for real-world FGIR applications.

\section{Conclusion}
\label{sec_conclusion}
Overall, our results show that stronger performance is generally accompanied by higher computational cost. Across backbones, we observe a non-uniform accuracy–efficiency relationship, with modern architectures such as Swin and ConvNeXt achieving the best trade-offs. Similarly, increasing input resolution consistently improves accuracy but incurs higher computational cost, with limited gains under constrained training settings such as FZ, indicating that naive resolution scaling is not an efficient strategy. For training and evaluation settings, more complex configurations consistently improve accuracy by up to 60\% in relative terms, while incurring up to a 279\% increase in computational cost, with additional gains observed from FGIR-specific methods such as CAL and CALMix. In particular, CAL improves accuracy significantly over FT but introduces higher computational cost in both training and inference. Building on this, CALMix further enhances performance, while our efficient variants CAL-NC and CALMix-NC remove inference-time cropping and maintain competitive accuracy with reduced computational overhead. Overall, these results highlight the effectiveness of FGIR-specific, data-aware training and demonstrate that carefully designed efficient variants can achieve a strong balance between accuracy and computational cost, providing practical guidance for future cost-aware FGIR system design.

{
\bibliographystyle{ieeenat_fullname}
    \bibliography{main}
}

\end{document}